\documentclass[10pt]{article} 
\usepackage[preprint]{tmlr}

\usepackage{amsmath,amsfonts,bm}

\newcommand{\ptrue}{P_{\mathrm{true}}}

\def\eqref#1{equation~\ref{#1}}

\def\1{\bm{1}}

\DeclareMathAlphabet{\mathsfit}{\encodingdefault}{\sfdefault}{m}{sl}
\SetMathAlphabet{\mathsfit}{bold}{\encodingdefault}{\sfdefault}{bx}{n}

\usepackage{hyperref}
\usepackage{url}
\usepackage{graphicx}
\usepackage{booktabs}
\usepackage{dsfont}
\usepackage{float}
\usepackage{mathtools}
\usepackage{tikz}
\usetikzlibrary{positioning, arrows.meta, shapes.multipart, calc}

\usepackage[english]{babel}
\addto\captionsenglish{%
}
\addto\extrasenglish{%
}

\title{Improved Confidence Estimates for Black-Box Large \\ Language Models}

\author{\name Sokhna Diarra Mbacke \email diarra@layer6.ai \\
      \addr Layer 6 AI
      \AND
      \name Mouloud Belbahri \email mouloud.belbahri@td.com \\
      \addr TD Insurance
      \AND
      \name Gabriel Loaiza-Ganem \email gabriel@layer6.ai\\
      \addr Layer 6 AI}

\def\month{MM}  
\def\year{YYYY} 
\def\openreview{\url{https://openreview.net/forum?id=XXXX}} 

\begin{document}

\maketitle

\begin{abstract}
  Uncertainty quantification (UQ) is essential for the safe deployment of large language models (LLMs). Existing methods, from verbalized confidence to ones requiring multiple generations, are often zero-shot and produce scores quantifying uncertainty without the need for labelled data. Nonetheless, in practice one must always evaluate their performance on a dataset of interest before deployment. In this work we show that, by leveraging this dataset, we consistently outperform these existing scores. Specifically, we build simple classifiers that predict LLM response correctness by using these scores and the correctness of similar queries as features. Our method produces minimal computational overhead, making it a cheap and straightforward enhancement for UQ in LLMs for real-world applications.
\end{abstract}

\section{Introduction}
The rapid advancement of Large Language Models (LLMs) has led to their integration across diverse sectors, from creative writing \citep{gomez2023confederacy} to automated code generation \citep{wang2021codet5, chen2021evaluating}, and more \citep{chaturvedi2023opportunities}. However, their susceptibility to hallucinations \citep{maynez2020faithfulness, xu2024hallucination} remains a significant barrier to deployment in high-stakes environments such as healthcare \citep{begoli2019need} and scientific research \citep{wang2023scientific}. Uncertainty Quantification (UQ) aims to address this challenge by evaluating model outputs to determine the degree to which a specific generation can be relied upon. Ultimately, establishing robust UQ frameworks is essential for unlocking the full potential of LLMs in many critical applications.

The field of UQ for LLMs is still largely open \citep{shorinwa2025survey}, and no clear consensus exists over what exactly should be quantified in the first place. For example, the traditional split of uncertainty into aleatoric and epistemic \citep{hullermeier2021aleatoric}---a staple of supervised learning---has been criticized as insufficient for the unique complexities of LLMs \citep{kirchhof2025position}. In practice, two major trends exist in the filed. The first, which we refer to as \emph{heuristic scoring}, produces abstract uncertainty scores designed solely to correlate with response correctness. These scores serve primarily as abstention tools, with the model withholding its response whenever the score fails to meet some predefined threshold. The second, \emph{confidence estimation}, attempts to directly estimate the probability of response correctness.

We argue that confidence should be the score of interest for UQ in LLMs: it subsumes heuristic scoring by maintaining the ability to trigger abstentions, while being more useful for downstream decision making. For example, calibrated probabilities of correctness can be integrated into decision-theoretical frameworks \citep{russel2010, berger2013statistical}, enabling systems to achieve optimal behaviour through the maximization of expected utility. 

In order to provide better confidence estimates than are currently available, we first observe that existing UQ methods for LLMs operate predominantly in a zero-shot manner, generating uncertainty scores without requiring a training dataset. In practice, however, models are rarely deployed without being evaluated on a representative labelled dataset. Our work begins with the insight that this data is a significant, yet untapped resource; by moving beyond the zero-shot paradigm, we can leverage this data to improve the reliability of LLM uncertainty estimates.

\begin{figure}
\centering
    \begin{tikzpicture}[
    scale=0.73,
    transform shape,
    >={Latex[length=2.5mm, width=1.5mm]},
    font=\sffamily,
    box/.style={
        draw=blue!40!black,
        fill=blue!10,
        rounded corners=2mm,
        align=center,
        minimum height=1.2cm,
        minimum width=2.4cm,
        thick,
        inner sep=2mm
    },
    orange box/.style={
        box,
        draw=orange!50!black,
        fill=orange!25
    },
    dataset/.style={
        rectangle split,
        rectangle split parts=4,
        draw=blue!40!black,
        fill=blue!10,
        thick,
        align=center,
        text width=3.4cm,
        inner sep=1.5mm,
        every node part/.style={align=center}
    }
]

\node[box] (q) {query \\ $q$};
\node[box, right=1.2cm of q] (r) {response(s) \\ $\mathbf{r}$};
\node[box, right=1.2cm of r] (s) {uncertainty scores \\ $\mathbf{s}$};
\node[box, right=1.2cm of s] (x) {features \\ $\mathbf{x}$};
\node[orange box, right=1.2cm of x] (f) {classifier \\ $f_\theta$};
\node[box, right=1.2cm of f] (y) {response correctness \\ $y$};

\node[dataset] (data) at ($(r.south)!0.5!(s.south) + (0,-2.5)$) {
    reference dataset
    \nodepart{two}
    $(q_1, \mathbf{r}_1, y_1)$
    \nodepart{three}
    $(q_2, \mathbf{r}_2, y_2)$
    \nodepart{four}
    $\dots$ \vphantom{$(q_1, r_1, y_1)$}
};

\foreach \source/\dest in {q/r, r/s, s/x, x/f, f/y} {
    \draw[->, thick, draw=black!80] (\source) -- (\dest);
}

\draw[->, thick, draw=black!80] (q.north) to[out=45, in=135, looseness=1.1] (s.north);

\draw[->, thick, draw=black!80] (q.south) to[out=-60, in=180] (data.west);
\draw[->, thick, draw=black!80] (data.east) to[out=0, in=-90] (x.south);

\end{tikzpicture}
\caption{\small{Most UQ methods use a given query $q$ and one or more corresponding LLM-generated responses $\mathbf{r}$ to obtain uncertainty scores $\mathbf{s}$ whose goal is to correlate with $y$ (a correctness label for the response). Instead, we retrieve similar queries from an existing dataset, and use the retrieved queries, alongside $\mathbf{s}$, to produce features $\mathbf{x}$; we then train a classifier predicting $y$ from $\mathbf{x}$.}}
    \label{fig:main}
    \vspace{-10pt}
\end{figure}
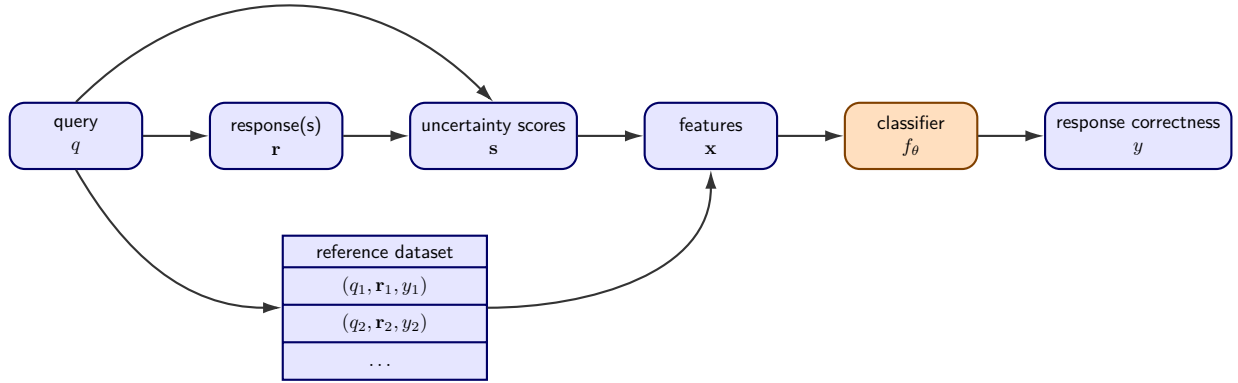

More specifically, we partition this dataset into a training set and a \emph{reference set}. To enhance any given uncertainty scores, we augment each query in the training data with these scores as features. We then leverage the reference set to engineer additional features: for every query in the training set, we identify its $k$-nearest neighbours within the reference set and extract neighbourhood-based statistics, such as the correctness of those neighbours and their cosine similarities to the query in embedding space. We then cast confidence estimation as a supervised binary classification task, leveraging the augmented training set to predict response correctness. We depict this process in \autoref{fig:main}.


Our framework offers several distinct advantages. First, the computational overhead is minimal: training the classifier is efficient due to its low complexity, and the marginal cost of inference is negligible relative to the cost of generating an LLM response. Second, the method requires only a dataset which, as previously noted, is a standard prerequisite in practical scenarios, even for zero-shot methods. Finally, our approach is inherently flexible; it provides a mechanism to aggregate several uncertainty scores into a single one, and it can transform heuristic scores into confidence estimates.

Empirically, our classifiers consistently outperform existing UQ methods across various datasets and models, and provide calibrated confidence estimates. Our results also reveal that while our data-driven approach consistently improves upon baseline uncertainty scores, the optimal classifier model is task-dependent. This variability underscores our central thesis: while labelled data is indeed a vital, untapped resource for UQ, leveraging it effectively necessitates the selection of a task-specific learner to account for different data distributions.

\section{Related Work and Motivation}\label{sec:related}
\paragraph{Heuristic scoring.} Although some heuristic scores are based on model logits \citep{duan2024shifting}, the majority rely on a sample-and-aggregate framework: multiple LLM responses are generated for a single query, and an uncertainty score is derived by summarizing the variability of these outputs into a scalar value. These methods diverge primarily in their mechanisms for generating responses and their specific summarization techniques. \citet{wang2023lora} use low-rank adaptation \citep[LoRA;][]{hu2022lora} to fine-tune an ensemble of LLMs to produce varied responses, and \citet{yang2024bayesian} use instead a Laplace approximation \hbox{\citep{ritter2018scalable}} on the LoRA blocks. \citet{hou2024decomposing} generate the responses by changing the query with different clarifications, and \citet{gao-etal-2024-spuq} with different perturbations. \citet{ross2025textual} treat the system prompt as a Bayesian parameter, and responses correspond to posterior samples. Other methods generate responses independently. \citet{kuhn2023semantic} group them into semantic clusters, and use the entropy of the corresponding distribution over clusters as the uncertainty score. \citet{lin2024generating} instead produce a matrix containing pairwise similarities for the generated responses, and produce various uncertainty scores based on the eigenspectra and traces of related matrices; \citet{nikitin2024kernel} use the corresponding von Neumann entropy instead. \citet{grewal2024improving} produce the score by averaging all the pairwise cosine similarities between generated responses, and \citet{qiu2024semantic} by using the density obtained from a kernel density estimator fitted on the embedding space of the responses.

\paragraph{Confidence estimation.} Verbalized confidence scores, often denoted as $\ptrue$, are obtained by directly prompting the LLM to assess the probability of its own response being correct \citep{kadavath2022language, yang2024verbalized}. However, these self-reported probabilities are prone to exhibit overconfidence \citep{zhou2024relying}. While \citet{kapoor2024calibration} propose a fine-tuning strategy to calibrate these verbalized outputs, this procedure is computationally expensive and requires white-box access to the LLM. Other works have explored the use of classifiers to predict response correctness, though with significant differences from our approach. For instance, \citet{mielke2022reducing} train the classifier using the internal representations of the LLM as features, which requires white-box access to the LLM. The closest work to ours is APRICOT \citep{ulmer2024calibrating}, where a classifier predicting response correctness is also trained. \citet{ulmer2024calibrating} use an auxiliary model to embed queries and responses, and use the resulting embeddings as features. Their framework remains distinct from ours in how features are constructed: it does not leverage a reference set, nor does it incorporate existing uncertainty scores. As a result, unlike our proposed method, it cannot function as a general-purpose tool for transforming heuristic scores into a confidence estimate.

\paragraph{Conformal methods.} Conformal prediction \citep{papadopoulos2002inductive, vovk2005algorithmic, angelopoulos2021gentle, angelopoulos2024theoretical} is a UQ method commonly used in supervised learning whose goal is to output a set of labels which is guaranteed to contain the ground truth label with some user-specified probability, with the size of the produced set quantifying uncertainty. Recent work \citep{quach2024conformal, kladny2025conformal, loaiza2026conf} has leveraged these ideas in the context of LLMs to produce sets of responses. These UQ methods require either a heuristic UQ score or a confidence estimate to obtain their sets; they are thus both orthogonal and compatible with our work.

\paragraph{Motivation.} Following the principle that uncertainty scores should be grounded in their usefulness for decision-making \citep{smith2025rethinking}, we observe that while current heuristic scores capture distinct types of uncertainty, they all share the fundamental goal of correlating with response correctness to guide actions like abstention. This observation motivates a framework that can integrate multiple heuristic scores into a single uncertainty metric. We contend that directly predicting the probability of correctness is the most effective way to achieve this; unlike heuristic scores, calibrated probabilities of the outcomes of interest---in this case correctness or lack thereof---provide a principled basis for optimal decision-making. Our first goal, therefore, is to develop a method that can produce a single confidence estimate from one or more pre-existing uncertainty scores, which also offers the additional benefit of being more interpretable than heuristic scores.

We establish two desiderata for our confidence estimates. First, our framework should not introduce requirements beyond those already imposed by its constituent uncertainty scores. For instance, if the pre-existing scores only require black-box access or a single model response, our method should maintain these constraints rather than necessitating white-box access or multiple generations. We will focus on this black-box setting since current state-of-the-art LLMs are all black-box, although our method could also be applied to white-box models. While many current uncertainty scores are zero-shot and require no labelled data, we have already argued that such data is available in any realistic scenario. Our second goal, therefore, is to leverage this available data to produce the best possible confidence estimates from existing scores.

\section{Background: Calibration}
Consider a target binary variable $Y$ and some features $X$. A classifier $f$ is said to be calibrated if, for every $p \in [0, 1]$ such that the event $f(X)=p$ is feasible,
\begin{equation}
    \mathbb{P}(Y=1|f(X)=p)=p.
\end{equation}
In words, calibration means that the output of the classifier can be trusted as a reliable probability estimate. Note that calibration is distinct from accuracy; for example a classifier might be accurate without being calibrated. Similarly, a calibrated classifier need not be accurate, e.g., the constant classifier $f(x)=\mathbb{P}(Y=1)$ is perfectly calibrated. Therefore, ideal classifiers should be both accurate and calibrated.

In practice, the calibration of a binary classifier is evaluated through its expected calibration error (ECE). Given a dataset ${(x_i, y_i)}_{i=1}^{n}$, the $[0,1]$ interval is partitioned into $L$ sub-intervals of equal length, $I_\ell$, which are then used to obtain data bins, $B_\ell \coloneqq \{(x_i, y_i) : f(x_i) \in I_\ell\}$. Then,
\begin{equation}
\mathrm{ECE}(f) \coloneqq \sum_{\ell} \frac{|B_\ell|}{n} \left| \text{acc}(f, B_\ell) - \text{conf}(f, B_\ell) \right|,
\end{equation}
where the sum is over all non-empty bins, $\text{acc}(f, B_\ell)$ is the accuracy of $f$ on bin $B_\ell$, and $\text{conf}(f, B_\ell)$ is its confidence on this bin, i.e., its average value.

Many binary classifiers produce a logit $h(x)$, so that $f$ is given by $f(x)=\sigma(h(x))$, where $\sigma$ denotes the sigmoid function. Temperature scaling is a popular strategy in deep learning \citep{guo2017calibration}, which modifies a pre-trained classifier by changing its output to $\sigma(h(x)/\tau)$, where $\tau > 0$ is fit to minimize ECE; importantly, the resulting classifier maintains the accuracy of $f$, but improves its calibration.

\section{Method}
\label{sec:methodology}



Rather than proposing a new uncertainty score from first principles, we ask a simpler question:
\begin{equation*}
    \parbox{15cm}{\emph{Given the outputs of existing uncertainty quantification methods and access to labelled evaluation data, what is the best estimate of the probability that an LLM response is correct?}}
\end{equation*}
Framing UQ in this way naturally leads to a supervised learning formulation, in which existing uncertainty scores and related statistics are treated as informative features, and the quantity of interest is the probability of correctness itself. We consider a fixed black-box LLM deployed on a task of interest and denote the space of text as $\mathcal{T}$. For a query (prompt) \(Q \in \mathcal{T}\), the LLM produces a response \(R \in \mathcal{T}\). We denote by $Y \in \{0,1\}$ the corresponding binary correctness label, i.e., $Y=1$ if and only if $R$ is a correct response to $Q$. For a fixed query-response pair, $(q, r)$, our goal is to estimate the probability of correctness,
\begin{equation}
    \eta(q, r) := \mathbb{P}\big(Y=1 \mid Q=q, R=r\big),
\end{equation}
which we refer to as the \emph{confidence} of the LLM response.

\paragraph{Available data.} As mentioned in \autoref{sec:related}, many heuristic UQ scores generate multiple responses $\mathbf{R} \coloneqq (R^{(1)}, \dots, R^{(m)}) \in \mathcal{T}^m$ for a single query $Q$. Here $Y \in \{0,1\}$ will correspond to the correctness of the first response, $R^{(1)}$, since we think of all the other responses as auxiliary for the computation of uncertainty scores. The way in which each response is generated can change depending on the underlying uncertainty score. For example, the first response could be generated with a low temperature, and subsequent responses could be generated \emph{i.i.d.}\ with a higher temperature (as in semantic entropy \citep{kuhn2023semantic}), or each one could instead correspond to a paraphrased version of $Q$. We also highlight that $m=1$ corresponds to the case where the underlying uncertainty scores require a single generation (e.g., $\ptrue$). We assume access to a labelled dataset,
\begin{equation}
    \mathcal{D} \coloneqq \{(q_i, \mathbf{r}_i, y_i)\}_{i=1}^n,
\end{equation}
containing queries $q_i \in \mathcal{T}$, responses $\mathbf{r}_i \in \mathcal{T}^m$, and their correctness label $y_i \in \{0,1\}$. 
As discussed earlier, such a dataset is typically available in practice, as it is required to assess the performance of UQ methods prior to deployment. We partition \(\mathcal{D}\) into two disjoint subsets. The first is a training set, denoted \(\mathcal{D}_{\mathrm{train}}\), which is used to fit the confidence model. The second is a reference set, denoted \(\mathcal{D}_{\mathrm{ref}}\), which is used solely to construct auxiliary features for new queries by summarizing the behavior of the LLM on similar, previously evaluated queries.
The reference set plays a distinct role from the training set and is central to our approach.

\paragraph{Baseline uncertainty scores as features.} A key novelty of our framework is to treat uncertainty scores as \emph{features} rather than final confidence. In doing so, we do not modify or replace existing UQ methods; instead, we reuse their outputs in a supervised learning formulation. We let
\begin{equation}
    \mathbf{S}(q,\mathbf{r}) \coloneqq \big(S_1(q,\mathbf{r}), \dots, S_c(q,\mathbf{r})\big) \in \mathbb{R}^c
\end{equation}
denote a collection of $c$ pre-existing uncertainty scores, $S_j(q, \mathbf{r}) \in \mathbb{R}$ for $j=1,\dots,c$, associated with a query \(q\) and its responses \(\mathbf{r}\). These may include heuristic scores or verbalized confidence estimates. 
An important consequence of this design is that our method introduces no additional requirements beyond those already imposed by the chosen scores. If the scores are computed from a single model generation ($m=1$), our method does not require additional sampling. This flexibility allows our framework to act as a general-purpose mechanism for aggregating and refining existing uncertainty estimates, rather than as a competing scoring method.


\paragraph{Reference set and neighbourhood-based features.} Existing uncertainty scores use only the information derived from a single query and its associated model outputs, without accounting for how the LLM has performed on similar queries in the past. However, when labelled evaluation data is available, this historical information provides a valuable empirical signal about response correctness that is not captured by standalone scores. 
To exploit this signal, we leverage the reference set \(\mathcal{D}_{\mathrm{ref}}\) and construct neighbourhood-based features that summarize the behavior of the LLM on queries similar to a given input.


\paragraph{Embedding space and similarity.} We begin by embedding queries into a shared representation space. Let $\phi : \mathcal{T} \to \mathbb{R}^d$ be an embedding function mapping queries to a vector space. For two queries \(q, q' \in \mathcal{T}\), we define their similarity as the cosine similarity between the embeddings,
\begin{equation}
    \mathrm{sim}(q, q') := 
\frac{\langle \phi(q), \phi(q') \rangle}
{\|\phi(q)\|_2 \, \|\phi(q')\|_2},
\end{equation}
where \(\langle \cdot, \cdot \rangle\) denotes the inner product and \(\|\cdot\|_2\) the corresponding \(\ell_2\)-norm. 
Now, for a query \(q\), let
\begin{equation}
    \mathcal{N}_k(q) \subset \mathcal{D}_{\mathrm{ref}}
\end{equation}
denote the set of its \(k\)-nearest neighbours in the reference set under the similarity measure defined above.

\paragraph{Neighbourhood statistics.} From \(\mathcal{N}_k(q)\), we construct summary features that capture both the LLM's past behavior on similar queries and the local structure of the query space. In particular, we consider:
\begin{itemize}
    \item The vector of neighbour correctness values, 
    \begin{equation}
        (y_i)_{i: q_i \in \mathcal{N}_k(q)},
    \end{equation}
    as well as the corresponding empirical mean and standard deviation,
    \begin{equation}
        \mu_y(q) \coloneqq \frac{1}{k} \sum_{i: q_i \in \mathcal{N}_k(q)} y_i \quad \text{and} \quad \sigma_y(q) \coloneqq \Big(\frac{1}{k-1} \sum_{i: q_i \in \mathcal{N}_k(q)} (y_i - \mu_y(q))^2\Big)^{1/2}.
    \end{equation}
    \item The vector of similarities to each neighbour,
    \begin{equation}
        (\mathrm{sim}(q, q_i))_{i: q_i \in \mathcal{N}_k(q)},
    \end{equation}
    as well as the corresponding empirical mean and standard deviation,
    \begin{equation}
        \mu_{\mathrm{sim}}(q) \coloneqq \frac{1}{k} \sum_{q_i \in \mathcal{N}_k(q)} \mathrm{sim}(q, q_i) \quad \text{and} \quad \sigma_{\mathrm{sim}}(q) \coloneqq \Big(\frac{1}{k-1} \sum_{q_i \in \mathcal{N}_k(q)} (\mathrm{sim}(q, q_i) - \mu_{\mathrm{sim}}(q))^2\Big)^{1/2}.
    \end{equation}
\end{itemize}




While the vectors of individual neighbour labels and similarities contain strictly more information, their empirical means (and/or standard-deviation) provide low-dimensional, low-variance summaries of local correctness and similarity structure. Including both aggregated and instance-level features allows downstream classifiers to trade off bias and variance, and to exploit neighbourhood information effectively across different model classes. We will denote the vector containing all these features---i.e., both instance level correctness labels and neighbour similarities, as well as their empirical means and standard deviations---as $\mathbf{f}_{\mathrm{ref}}(q) \in \mathbb{R}^{2k+4}$.

The use of a reference set in this way allows us to reuse past labelled queries to inform uncertainty estimates for new queries. Intuitively, these neighbourhood-based features provide a local estimate of the LLM’s reliability: if past queries similar to \(q\) tended to be answered correctly, it is more likely that the current response is also correct. Likewise, the distances to neighbours reflect how representative the reference points are for the current query. By summarizing both correctness and proximity information, the model gains access to empirical patterns that are not captured by standalone uncertainty scores, making it easier to predict response correctness without altering the LLM itself.

\paragraph{Feature representation.} For a given query-responses pair $(q, \mathbf{r})$ we construct the feature vector
\begin{equation}
    \mathbf{X}(q, \mathbf{r}) \coloneqq \big(\mathbf{S}(q,\mathbf{r}), \mathbf{f}_{\mathrm{ref}}(q)\big) \in \mathbb{R}^{c+2k+4}.
\end{equation}
This framework is modular: features can be added or removed depending on availability and constraints. In the absence of a reference set, our method reduces to a supervised aggregation of existing uncertainty scores. We also experimented with augmenting the feature representation using high-dimensional query embeddings, in the spirit of approaches such as the one from \citet{ulmer2024calibrating}. However, we did not observe consistent improvements over models based on uncertainty scores and neighbourhood statistics alone. We conjecture that this is due to the high dimensionality of embedding representations, which may require substantially more complex models and larger training sets to reliably extract signals predictive of correctness.

\paragraph{Confidence estimation via supervised learning.} We cast confidence estimation as a supervised binary classification problem. Let $f_\theta : \mathbb{R}^{c+2k+4} \to [0,1]$ be a probabilistic classifier parameterized by \(\theta\). We interpret
\begin{equation}
    \hat{\eta}(q, r^{(1)}) := f_\theta\big(\mathbf{X}(q,\mathbf{r})\big)
\end{equation}
as an estimate of $\eta(q, r^{(1)}) = \mathbb{P}(Y=1 \mid Q=q, R^{(1)}=r^{(1)})$. 
The classifier is trained on \(\mathcal{D}_{\mathrm{train}}\) by minimizing a proper scoring rule such as the cross entropy loss,
\begin{equation}
    \dfrac{1}{|\mathcal{D}_{\mathrm{train}}|} \sum_{(q_i, \mathbf{r}_i, y_i) \in \mathcal{D}_{\mathrm{train}}}
    \big[
- y_i \log f_\theta(\mathbf{x}_i)
- (1-y_i)\log(1-f_\theta(\mathbf{x}_i))
\big],
\end{equation}
where \(\mathbf{x}_i=\mathbf{X}(q_i,\mathbf{r}_i)\) is the $i$-th observed feature vector from the training set. 

Our approach improves calibration by explicitly learning the conditional probability of correctness given informative features, rather than relying on heuristic scores alone. Even when $\mathbf{S}(q,\mathbf{r})$ is a confidence estimate (e.g., $\ptrue$), it is generally not calibrated, e.g., LLM are often over-confident in their responses. By contrast, when a probabilistic classifier is trained using a proper scoring rule, the Bayes-optimal predictor satisfies
\begin{equation}
    f^*(\mathbf{x}) = \mathbb{P}(Y=1 \mid \mathbf{X}=\mathbf{x}).
\end{equation}
Because our feature representation includes the baseline scores and additional neighbourhood-based statistics, the learned predictor conditions on strictly more information that if it was just given the scores. Under mild regularity conditions, conditioning on additional informative variables cannot worsen calibration and can strictly improve it when the added features are predictive of correctness \citep{murphy1973new,dawid1982well,gneiting2007strictly}.

Intuitively, the classifier learns how baseline scores should be rescaled and combined for the task at hand, while neighbourhood features provide empirical estimates of local correctness frequencies. As a result, the learned confidence estimates converge toward calibrated probabilities on the data distribution, whereas fixed heuristic scores cannot adapt to dataset-specific biases.

\paragraph{Inference and computational considerations.} At inference time, for a given query-responses pair $(q, \mathbf{r})$, producing $\hat{\eta}(q, r^{(1)})$ proceeds in three steps: $(i)$ computing the baseline uncertainty scores \(\mathbf{S}(q, \mathbf{r})\); $(ii)$ retrieving the \(k\)-nearest neighbours from the reference set \(\mathcal{D}_{\mathrm{ref}}\); and $(iii)$ evaluating the trained classifier \(f_\theta\) on the combined feature vector. 
For step $(i)$, note that any computational requirements imposed by a score \(S_j\) (e.g., multiple generations) are inherited by our method and not introduced by it. Steps $(ii)$ and $(iii)$, as well as training the classifier, incur negligible computational overhead relative to generating LLM responses, meaning that the computational cost of our method is essentially the same as that of the underlying baseline scores.




\paragraph{Auto-ML for classifier selection.} 

As a minimal baseline, we first consider a temperature-scaled version of the verbalized confidence score (ts-$\ptrue$), which can be viewed as the simplest post-hoc model for producing calibrated confidence estimates. Specifically, this approach applies a monotone rescaling to the self-reported probability output by the LLM, requiring only a single scalar parameter and no additional features. It represents the least expressive model one can fit while still leveraging labeled data to improve calibration. Beyond this baseline, rather than training a single classifier, we adopt an Auto-ML approach to select the classifier \(f_\theta\) that performs best for a given task and feature set. All candidate models are trained on the same feature representations and compared using a unified validation protocol. 

In our experiments, we consider two classes of probabilistic classifiers. First, we train an \(\ell_1\)-regularized logistic regression model, which estimates the probability of correctness. The regularization parameter is selected via \(k\)-fold cross-validation by maximizing the area under the receiver operating characteristic curve (AUROC). Second, we train a random forest classifier \citep{breiman2001random} to capture nonlinear interactions between features, although the framework is fully compatible with other classifiers, including multi-layer perceptrons, gradient-boosted trees, or any off-the-shelf supervised learner. For random forests, we perform a grid search over the number of trees, maximum tree depth, and the minimum number of samples required to split an internal node. Hyperparameters are selected using \(k\)-fold cross-validation, again optimizing AUROC. We additionally apply temperature scaling as a post-hoc calibration step, fitting a single scalar temperature on held-out data to rescale the predictions while preserving their ordering.

Once all the models are trained, we select the one having achieved the best AUROC. While the feature construction process remains fixed across tasks, the resulting estimator \(f_\theta\) can vary substantially depending on the task, dataset, and feature representation. Linear models may suffice when baseline uncertainty scores are already well aligned with correctness, whereas nonlinear models can better exploit complex interactions induced by neighbourhood-based information. The Auto-ML procedure enables task-adaptive confidence estimation without modifying the LLM or the underlying uncertainty scores.




\section{Experiments}\label{sec:experiments}
\newcommand{\pptrue}{$\ptrue$}
\newcommand{\ptruepred}{$\ptrue$-pred}
\newcommand{\ptruethresh}{$\ptrue$-thresh}
\newcommand{\tempscaledptrue}{ts-$\ptrue$}

In this section, we present empirical results demonstrating that our framework consistently improves upon existing UQ scores. We evaluate our method across multiple datasets, training simple classifiers—logistic regression and random forests—on feature sets that include baseline UQ scores, verbalized confidence estimates, and neighbourhood-based correctness/distance statistics. The objective in all experiments is to predict the correctness of the LLM response. 
Our code is available at \url{https://github.com/layer6ai-labs/improved_llm_confidence}.

\subsection{Experimental Setup}

\paragraph{Datasets.} We consider the following datasets: CommonSense QA (CS\_QA) \citep{common_sense_qa}, Natural Questions (NQ) \citep{nq}, SciQ \citep{sciq}, and SimpleQA \citep{simpleqa}. NQ contains real Google search queries with Wikipedia-sourced answers; we randomly sampled 10,000 datapoints from the training split. We also randomly sampled 8,000 queries each from the training splits of SciQ, a 4-option multiple-choice science exam dataset, and CS\_QA, a 5-option multiple-choice reasoning benchmark. For SimpleQA, a dataset from OpenAI that measures short-form factuality, we kept the full dataset of 4,326 examples.

\paragraph{LLM models.} To evaluate our method, we worked primarily with the GPT4.1 model family, specifically GPT4.1 (flagship), GPT4.1-mini, and GPT4.1-nano. This choice was guided by two factors. $(i)$ Practical applicability: as closed-source black-box models, these represent the most common type of LLM encountered in real-world production environments where internal weights are inaccessible. $(ii)$ Scalability analysis: by testing across three distinct sizes (standard, mini, and nano), we can observe how our method’s performance and reliability scale with model capacity. Although our main objective is to perform UQ for black-box models, we also consider LLaMA4Maverick, an open-source model, as a way to highlight the broad applicability of our procedure.

\paragraph{Evaluation.} 
Below we summarize the heuristic scores that we build upon and compare against, please refer to \autoref{app:baselines} for more details. All methods are evaluated on a shared test set across five random seeds, with performance assessed using average AUROC and ECE.


\textbf{Confidence UQ scores.} We evaluate two variants of the verbalized confidence score (\(\ptrue\)): the raw \(\ptrue\) used directly as a probabilistic predictor, and a temperature-scaled version (\tempscaledptrue) that refines calibration via a single learned scalar. We also evaluate the APRICOT method, and used the default configuration in the code of \citet{ulmer2024calibrating} for training the neural network predicting confidence.

\textbf{Semantic-based heuristic UQ scores.} We compare against three semantic uncertainty methods: semantic entropy \citep{kuhn2023semantic}, Laplacian \citep{lin2024generating}, and kernel entropy \citep{nikitin2024kernel}. These methods follow a common procedure: for each query, we sample a total of $m=21$ responses; the first response corresponds to the response whose uncertainty we wish to quantify and it is sampled with greedy decoding, whereas the other $20$ responses are sampled at a high temperature ($T=1$) to induce higher diversity in the generated responses so as to obtain more meaningful UQ scores. For each model and dataset, we report the best performance achieved by these three methods.

\textbf{Model-based.} We evaluate our approach following the Auto-ML framework described in \autoref{sec:methodology}. Concretely, we train \(\ell_1\)-regularized logistic regression and random forest models, reporting the best performance achieved between the two. Neighbourhood-based features are constructed using \(k=20\) nearest neighbours from the reference set. Details on the construction of the reference set using determinantal point processes (DPPs) \citep{kulesza2011kdpp, kulesza2012determinantal} are provided in \autoref{sec:dpp}.

\begin{table}[t]
\centering
\footnotesize
\caption{AUROC (\textbf{higher is better}) and ECE (\textbf{lower is better}) across methods. For each method family, we report the best score (highest AUROC, lowest ECE), averaged over five independent runs with different random seeds. Semantic UQ scores are included as features in the Auto-ML method.}
\begin{tabular}{l ccccc cccc}
\toprule
 & \multicolumn{4}{c}{AUROC ($\uparrow$)} & \multicolumn{4}{c}{ECE ($\downarrow$)} \\
\cmidrule(lr){2-5} \cmidrule(l){6-9}
 & $\ptrue$ & Semantic & APRICOT & Auto-ML & $\ptrue$ & ts-$\ptrue$ & APRICOT & Auto-ML \\
\midrule
\multicolumn{9}{l}{\textbf{GPT4.1}} \\
\midrule
CS\_QA  & 0.743 & 0.645 & 0.494 & \textbf{0.770} & 0.069 & \textbf{0.023} & 0.660 & 0.024 \\
NQ  & 0.715 & 0.709 & 0.457 & \textbf{0.774} & 0.322 & \textbf{0.016} & 0.148 & 0.029 \\
SciQ  & 0.895 & 0.653 & 0.452 & \textbf{0.911} & 0.021 & \textbf{0.008} & 0.966 & \textbf{0.008} \\
SimpleQA  & 0.632 & 0.769 & 0.489 & \textbf{0.782} & 0.325 & 0.183 & 0.312 & \textbf{0.040} \\
\midrule
\multicolumn{9}{l}{\textbf{GPT4.1-mini}} \\
\midrule
CS\_QA  & 0.697 & 0.648 & 0.498 & \textbf{0.741} & 0.085 & 0.081 & 0.584 & \textbf{0.035} \\
NQ  & 0.703 & 0.725 & 0.504 & \textbf{0.762} & 0.330 & 0.015 & \textbf{0.011} & 0.028 \\
SciQ  & 0.837 & 0.702 & 0.456 & \textbf{0.874} & 0.041 & \textbf{0.005} & 0.959 & 0.007 \\
SimpleQA  & 0.640 & \textbf{0.790} & 0.412 & 0.786 & 0.113 & 0.111 & 0.766 & \textbf{0.024} \\
\midrule
\multicolumn{9}{l}{\textbf{GPT4.1-nano}} \\
\midrule
CS\_QA  & 0.626 & 0.693 & 0.508 & \textbf{0.710} & 0.106 & 0.027 & 0.462 & \textbf{0.024} \\
NQ  & 0.734 & 0.751 & 0.496 & \textbf{0.790} & 0.304 & 0.129 & 0.221 & \textbf{0.032} \\
SciQ  & 0.769 & 0.719 & 0.543 & \textbf{0.843} & 0.034 & \textbf{0.008} & 0.915 & 0.013 \\
SimpleQA  & 0.644 & 0.725 & 0.490 & \textbf{0.755} & 0.056 & 0.056 & 0.943 & \textbf{0.007} \\
\bottomrule
\end{tabular}
\label{tab-sem-metrics}
\end{table}

\subsection{Results}

Overall, our experimental results demonstrate that leveraging existing uncertainty scores and neighbourhood statistics as features in a simple classifier substantially improves predictive accuracy and yields well-calibrated confidence estimates for LLM responses. This is shown in \autoref{tab-sem-metrics}, which compares our Auto-ML method when using semantic UQ scores and neighbourhood statistics as features against various baselines. Our proposed Auto-ML procedure consistently outperforms all the baselines in AUROC---there is a single instance where it does not, and the difference is extremely minor. We also note that APRICOT performs quite poorly. Although we could likely improve this baseline by tuning it separately for each task, we highlight that we carried out a fair comparison, as our method involves no tuning. Indeed, avoiding having to tune hyperparameters for the confidence predictor is an important advantage of our method.

\autoref{tab-sem-metrics} also reports the best ECE across method families under the same setting; note that we do not compare ECE against semantic methods since they do not provide a confidence score. Our Auto-ML approach consistently beats $\ptrue$, and while it does not always beat ts-$\ptrue$, it is often either a close second or a distant first. We also highlight that ts-$\ptrue$ is in a sense the simplest model we can train using an existing score to predict confidence, and so it can also be viewed as an instantiation of our proposed methodology. Overall, \autoref{tab-sem-metrics} shows that the confidence scores provided by our method are both highly predictive of correctness and well calibrated. We also emphasize once again that the computational overhead of our Auto-ML confidence over the ``Semantic'' baseline is negligible.

\begin{figure*}[t]
    \centering
    \includegraphics[width=\textwidth]{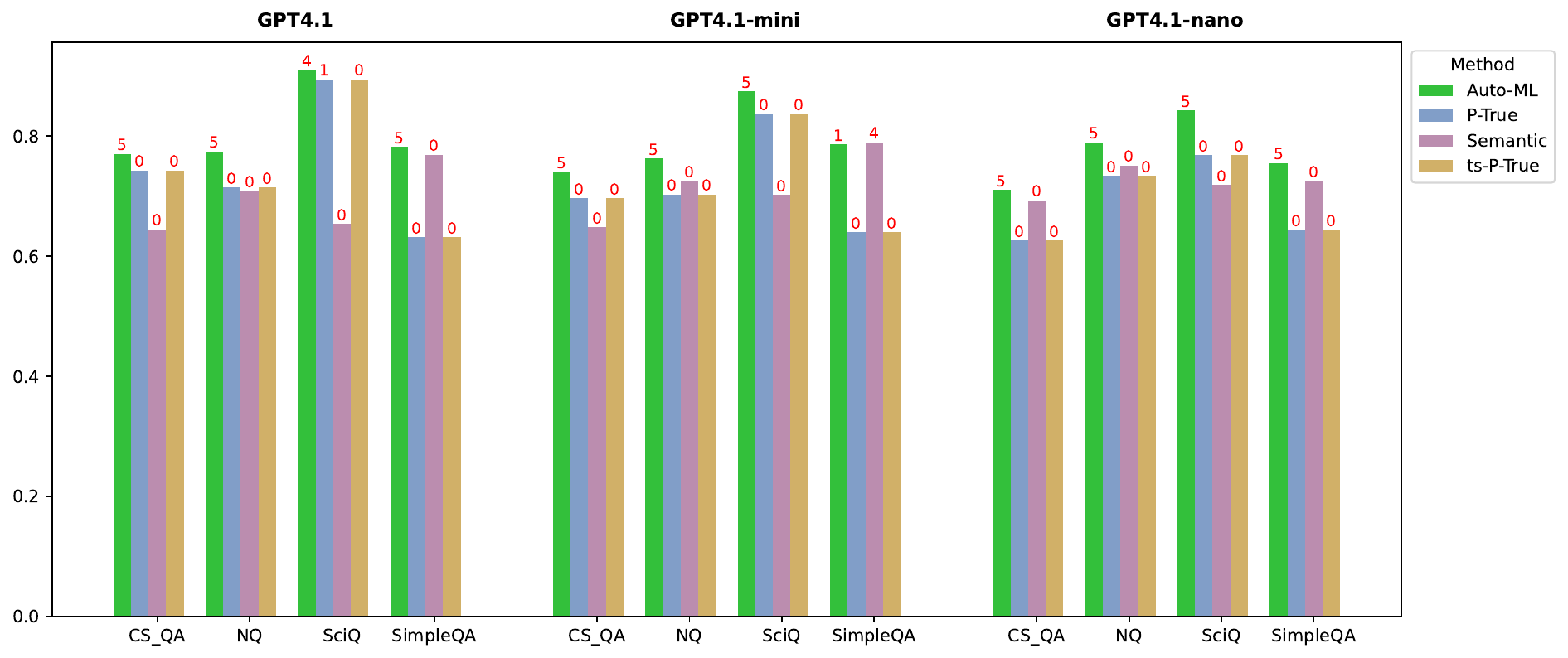}
    \caption{AUROC (\textbf{higher is better}) across methods. Each bar shows the average AUROC over five independent runs with different random seeds for each dataset/LLM combination. Red integers above the bars indicate "win" counts, i.e., the number of times a method achieved the highest AUROC among all methods. Semantic UQ scores are included as features in the Auto-ML method.}
    \label{fig-auroc_with_sem}
\end{figure*}

\autoref{fig-auroc_with_sem} shows AUROC results and ``win" counts across correctness prediction methods for all scenarios. Our Auto-ML method consistently outperforms the baselines. Interestingly, on multiple-choice datasets, semantic entropy-based methods perform poorly, being outperformed by both the Auto-ML method and simple verbalized confidence (\pptrue). More detailed results are provided in \autoref{app:add_results}.

\begin{table*}[t]
\centering
\caption{AUROC (\textbf{higher is better}) and ECE (\textbf{lower is better}) across methods with LLaMA4Maverick generations.}
\begin{tabular}{l ccc cc}
\toprule
 & \multicolumn{3}{c}{AUROC ($\uparrow$)} & \multicolumn{2}{c}{ECE ($\downarrow$)} \\
\cmidrule(lr){2-4} \cmidrule(l){5-6}
 & \pptrue & Semantic & Auto-ML & \pptrue & Auto-ML \\
\midrule
\multicolumn{6}{l}{\textbf{LLaMA4Maverick}} \\
\midrule
NQ & 0.727 & 0.762 & \textbf{0.796} & 0.077 & \textbf{0.029} \\
SciQ & 0.853 & 0.580 & \textbf{0.857} & \textbf{0.003} & \textbf{0.003} \\
SimpleQA & 0.624 & 0.778 & \textbf{0.787} & 0.336 & \textbf{0.027} \\
\bottomrule
\end{tabular}
\label{tab:llama4maverick-metrics}
\end{table*}

\paragraph{Ablation: changing the model family.} 
\autoref{tab:llama4maverick-metrics} shows analogous comparisons as those in \autoref{tab-sem-metrics} for LLaMA4Maverick: we can see that our Auto-ML method still provides improved UQ scores.

\paragraph{Ablation: removing semantic features.} 
We also evaluate in \autoref{tab-no-sem-metrics} our method using only verbalized confidence (\pptrue) and neighbourhood-based statistics as features, omitting semantic UQ metrics; this requires generating only a single response ($m=1$) and thus has comparable computational cost to $\ptrue$. Even in this restricted setting, our method achieves strong performance. 
These results highlight the efficiency of the Auto-ML approach: using only the greedy generation, the verbalized confidence, and a reference set (readily available in practice), the model leverages neighborhood information to produce well calibrated predictions. While this single-sample strategy cannot consistently outperform semantic UQ methods---given the disparity in computational effort and available information---it still delivers competitive AUROC and ECE scores.

\begin{table*}[t]
\centering
\caption{AUROC (\textbf{higher is better}) and ECE (\textbf{lower is better}) across methods \textbf{without} semantic features for Auto-ML.}
\begin{tabular}{l cccc ccc}
\toprule
 & \multicolumn{3}{c}{AUROC ($\uparrow$)} & \multicolumn{3}{c}{ECE ($\downarrow$)} \\
\cmidrule(lr){2-4} \cmidrule(l){5-7}
 & $\ptrue$  & Semantic & Auto-ML & $\ptrue$ & ts-$\ptrue$ & Auto-ML \\
\midrule
\multicolumn{7}{l}{\textbf{GPT4.1}} \\
\midrule
CS\_QA & 0.743 & 0.645 & \textbf{0.747} & 0.069 & \textbf{0.023} & 0.036 \\
NQ & 0.715 & 0.709 & \textbf{0.735} & 0.322 & \textbf{0.016} & 0.025 \\
SciQ  & \textbf{0.895} & 0.653 & 0.894 & 0.021 & \textbf{0.008} & 0.009 \\
SimpleQA & 0.632 & \textbf{0.769} & 0.667 & 0.325 & 0.183 & \textbf{0.088} \\
\midrule
\multicolumn{7}{l}{\textbf{GPT4.1-mini}} \\
\midrule
CS\_QA & 0.697 & 0.648 & \textbf{0.711} & 0.085 & 0.081 & \textbf{0.046} \\
NQ & 0.703 & \textbf{0.725} & 0.721 & 0.330 & \textbf{0.015} & 0.062 \\
SciQ & 0.837 & 0.702 & \textbf{0.858} & 0.041 & \textbf{0.005} & 0.010 \\
SimpleQA & 0.640 & \textbf{0.790} & 0.649 & 0.113 & 0.111 & \textbf{0.076} \\
\midrule
\multicolumn{7}{l}{\textbf{GPT4.1-nano}} \\
\midrule
CS\_QA & 0.626 & \textbf{0.693} & 0.638 & 0.106 & \textbf{0.027} & 0.031 \\
NQ & 0.734 & 0.751 & \textbf{0.753} & 0.304 & 0.129 & \textbf{0.114} \\
SciQ & 0.769 & 0.719 & \textbf{0.778} & 0.034 & \textbf{0.008} & 0.012 \\
SimpleQA & 0.644 & \textbf{0.725} & 0.662 & 0.056 & 0.056 & \textbf{0.033} \\
\bottomrule
\end{tabular}
\label{tab-no-sem-metrics}
\end{table*}

\subsection{Discussion}
\paragraph{Use of the semantic features.} The experimental results in \autoref{tab-sem-metrics} as well as in \autoref{fig-auroc_with_sem} show that by leveraging existing UQ methods, we can obtain an accurate and well calibrated predictor of correctness, without additional LLM calls, or access to the LLM's internal representations. Moreover, in our ablation experiments, we show that: $(i)$ even without the use of semantic features, we can still get a competitive predictor with much lower cost than the semantic methods; and that $(ii)$ our method performs well beyond the black-box GPT family.



\paragraph{Model choice.} We use logistic regression and random forests as our classification models due to their low computational overhead and inherent interpretability. These models serve as efficient tools for isolating signals within the feature space and establishing robust performance baselines. While our current implementation focuses on these simpler architectures, the framework is modular. Hence, higher-capacity predictors, such as deep neural networks, could be integrated to potentially capture more intricate, non-linear dependencies in the data.

\paragraph{Limitations.} Our approach has two primary limitations. First, it requires training dataset-specific classifiers, which may limit immediate out-of-the-box generalization. However, this is largely mitigated by the fact that most practical applications involve optimizing a model for a specific target domain or task. Second, the method's predictive accuracy depends on the reference set, as it requires neighbours for each question to extract meaningful local information. Consequently, a sufficiently large dataset is needed to provide high-quality retrieval. In practice, this requirement is typically satisfied during the standard model development lifecycle, where a representative dataset is already a prerequisite for evaluation and fine-tuning.

\section{Conclusions and Future Work}
In this work, we have demonstrated that confidence prediction for large language models should be treated as a machine learning problem. Our central thesis is that regardless of the initial UQ score, performance can almost always be improved by utilizing those scores as features within a supervised framework. While we used straightforward models like logistic regression and random forests to establish a baseline, the integration of more sophisticated Auto-ML pipelines is bound to further refine these predictions and yield even higher calibration and discriminative power.

Our advocacy for this classification-based approach is not a dismissal of research into heuristic UQ scores. On the contrary, we view heuristic scores as essential inputs, and our framework provides a systematic way of enhancing them by learning how to weigh and combine different signals effectively. Finally, while this paper focused on binary correctness, future research should expand this paradigm toward multi-class classification. Moving beyond a simple ``correct / incorrect'' binary allows for a more nuanced evaluation of model outputs by distinguishing between responses that are outright incorrect, partially correct, acceptable, or excellent, thereby providing a more robust foundation for reliable deployment LLMs.







\bibliography{bib}

\begin{thebibliography}{54}
\providecommand{\natexlab}[1]{#1}
\providecommand{\url}[1]{\texttt{#1}}
\expandafter\ifx\csname urlstyle\endcsname\relax
  \providecommand{\doi}[1]{doi: #1}\else
  \providecommand{\doi}{doi: \begingroup \urlstyle{rm}\Url}\fi

\bibitem[Angelopoulos \& Bates(2021)Angelopoulos and Bates]{angelopoulos2021gentle}
Anastasios~N Angelopoulos and Stephen Bates.
\newblock A gentle introduction to conformal prediction and distribution-free uncertainty quantification.
\newblock \emph{arXiv:2107.07511}, 2021.

\bibitem[Angelopoulos et~al.(2024)Angelopoulos, Barber, and Bates]{angelopoulos2024theoretical}
Anastasios~N Angelopoulos, Rina~Foygel Barber, and Stephen Bates.
\newblock Theoretical foundations of conformal prediction.
\newblock \emph{arXiv:2411.11824}, 2024.

\bibitem[Begoli et~al.(2019)Begoli, Bhattacharya, and Kusnezov]{begoli2019need}
Edmon Begoli, Tanmoy Bhattacharya, and Dimitri Kusnezov.
\newblock The need for uncertainty quantification in machine-assisted medical decision making.
\newblock In \emph{Nature Machine Intelligence}, volume~1, pp.\  20--23, 2019.

\bibitem[Berger(2013)]{berger2013statistical}
James~O Berger.
\newblock \emph{Statistical decision theory and {B}ayesian analysis}.
\newblock Springer Science \& Business Media, 2013.

\bibitem[Breiman(2001)]{breiman2001random}
Leo Breiman.
\newblock Random forests.
\newblock In \emph{Machine learning}, volume~45, pp.\  5--32. Springer, 2001.

\bibitem[Chaturvedi \& Verma(2023)Chaturvedi and Verma]{chaturvedi2023opportunities}
Rijul Chaturvedi and Sanjeev Verma.
\newblock \emph{Opportunities and Challenges of {AI}-Driven Customer Service}, pp.\  33--71.
\newblock Springer International Publishing, 2023.

\bibitem[Chen et~al.(2018)Chen, Zhang, and Zhou]{chen2018fast}
Laming Chen, Guoxin Zhang, and Eric Zhou.
\newblock Fast greedy map inference for determinantal point process to improve recommendation diversity.
\newblock \emph{Advances in Neural Information Processing Systems}, 31, 2018.

\bibitem[Chen et~al.(2021)Chen, Tworek, Jun, Yuan, de~Oliveira~Pinto, Kaplan, Edwards, Burda, Joseph, Brockman, Ray, Puri, Krueger, Petrov, Khlaaf, Sastry, Mishkin, Chan, Gray, Ryder, Pavlov, Power, Kaiser, Bavarian, Winter, Tillet, Such, Cummings, Plappert, Chantzis, Barnes, Herbert-Voss, Guss, Nichol, Paino, Tezak, Tang, Babuschkin, Balaji, Jain, Saunders, Hesse, Carr, Leike, Achiam, Misra, Morikawa, Radford, Knight, Brundage, Murati, Mayer, Welinder, McGrew, Amodei, McCandlish, Sutskever, and Zaremba]{chen2021evaluating}
Mark Chen, Jerry Tworek, Heewoo Jun, Qiming Yuan, Henrique~Ponde de~Oliveira~Pinto, Jared Kaplan, Harri Edwards, Yuri Burda, Nicholas Joseph, Greg Brockman, Alex Ray, Raul Puri, Gretchen Krueger, Michael Petrov, Heidy Khlaaf, Girish Sastry, Pamela Mishkin, Brooke Chan, Scott Gray, Nick Ryder, Mikhail Pavlov, Alethea Power, Lukasz Kaiser, Mohammad Bavarian, Clemens Winter, Philippe Tillet, Felipe~Petroski Such, Dave Cummings, Matthias Plappert, Fotios Chantzis, Elizabeth Barnes, Ariel Herbert-Voss, William~Hebgen Guss, Alex Nichol, Alex Paino, Nikolas Tezak, Jie Tang, Igor Babuschkin, Suchir Balaji, Shantanu Jain, William Saunders, Christopher Hesse, Andrew~N. Carr, Jan Leike, Josh Achiam, Vedant Misra, Evan Morikawa, Alec Radford, Matthew Knight, Miles Brundage, Mira Murati, Katie Mayer, Peter Welinder, Bob McGrew, Dario Amodei, Sam McCandlish, Ilya Sutskever, and Wojciech Zaremba.
\newblock Evaluating large language models trained on code.
\newblock \emph{arXiv:2107.03374}, 2021.

\bibitem[Dawid(1982)]{dawid1982well}
A~Philip Dawid.
\newblock The well-calibrated bayesian.
\newblock \emph{Journal of the American statistical Association}, 77\penalty0 (379):\penalty0 605--610, 1982.

\bibitem[Duan et~al.(2024)Duan, Cheng, Wang, Zavalny, Wang, Xu, Kailkhura, and Xu]{duan2024shifting}
Jinhao Duan, Hao Cheng, Shiqi Wang, Alex Zavalny, Chenan Wang, Renjing Xu, Bhavya Kailkhura, and Kaidi Xu.
\newblock Shifting attention to relevance: Towards the predictive uncertainty quantification of free-form large language models.
\newblock In \emph{Proceedings of the 62nd Annual Meeting of the Association for Computational Linguistics}, pp.\  5050--5063, 2024.

\bibitem[Gao et~al.(2024)Gao, Zhang, Mouatadid, and Das]{gao-etal-2024-spuq}
Xiang Gao, Jiaxin Zhang, Lalla Mouatadid, and Kamalika Das.
\newblock {SPUQ}: Perturbation-based uncertainty quantification for large language models.
\newblock In \emph{Conference of the European Chapter of the Association for Computational Linguistics}, 2024.

\bibitem[Gneiting \& Raftery(2007)Gneiting and Raftery]{gneiting2007strictly}
Tilmann Gneiting and Adrian~E Raftery.
\newblock Strictly proper scoring rules, prediction, and estimation.
\newblock \emph{Journal of the American statistical Association}, 102\penalty0 (477):\penalty0 359--378, 2007.

\bibitem[G{\'o}mez-Rodr{\'\i}guez \& Williams(2023)G{\'o}mez-Rodr{\'\i}guez and Williams]{gomez2023confederacy}
Carlos G{\'o}mez-Rodr{\'\i}guez and Paul Williams.
\newblock {A Confederacy of Models: a Comprehensive Evaluation of {LLM}s on Creative Writing}.
\newblock In \emph{Findings of the Association for Computational Linguistics: EMNLP 2023}, pp.\  14504--14528, 2023.

\bibitem[Grewal et~al.(2024)Grewal, Bonilla, and Bui]{grewal2024improving}
Yashvir~S Grewal, Edwin~V Bonilla, and Thang~D Bui.
\newblock Improving uncertainty quantification in large language models via semantic embeddings.
\newblock \emph{arXiv:2410.22685}, 2024.

\bibitem[Guo et~al.(2017)Guo, Pleiss, Sun, and Weinberger]{guo2017calibration}
Chuan Guo, Geoff Pleiss, Yu~Sun, and Kilian~Q Weinberger.
\newblock On calibration of modern neural networks.
\newblock In \emph{International conference on machine learning}, 2017.

\bibitem[He et~al.(2021)He, Liu, Gao, and Chen]{he2021deberta}
Pengcheng He, Xiaodong Liu, Jianfeng Gao, and Weizhu Chen.
\newblock Deberta: Decoding-enhanced bert with disentangled attention.
\newblock In \emph{International Conference on Learning Representations}, 2021.

\bibitem[Hou et~al.(2024)Hou, Liu, Qian, Andreas, Chang, and Zhang]{hou2024decomposing}
Bairu Hou, Yujian Liu, Kaizhi Qian, Jacob Andreas, Shiyu Chang, and Yang Zhang.
\newblock Decomposing uncertainty for large language models through input clarification ensembling.
\newblock In \emph{International Conference on Machine Learning}, 2024.

\bibitem[Hu et~al.(2022)Hu, Shen, Wallis, Allen-Zhu, Li, Wang, Wang, and Chen]{hu2022lora}
Edward~J Hu, Yelong Shen, Phillip Wallis, Zeyuan Allen-Zhu, Yuanzhi Li, Shean Wang, Lu~Wang, and Weizhu Chen.
\newblock Lo{RA}: Low-rank adaptation of large language models.
\newblock In \emph{International Conference on Learning Representations}, 2022.

\bibitem[H{\"u}llermeier \& Waegeman(2021)H{\"u}llermeier and Waegeman]{hullermeier2021aleatoric}
Eyke H{\"u}llermeier and Willem Waegeman.
\newblock Aleatoric and epistemic uncertainty in machine learning: An introduction to concepts and methods.
\newblock \emph{Machine learning}, 110\penalty0 (3):\penalty0 457--506, 2021.

\bibitem[Kadavath et~al.(2022)Kadavath, Conerly, Askell, Henighan, Drain, Perez, Schiefer, Hatfield-Dodds, DasSarma, Tran-Johnson, et~al.]{kadavath2022language}
Saurav Kadavath, Tom Conerly, Amanda Askell, Tom Henighan, Dawn Drain, Ethan Perez, Nicholas Schiefer, Zac Hatfield-Dodds, Nova DasSarma, Eli Tran-Johnson, et~al.
\newblock Language models (mostly) know what they know.
\newblock \emph{arXiv:2207.05221}, 2022.

\bibitem[Kapoor et~al.(2024)Kapoor, Gruver, Roberts, Pal, Dooley, Goldblum, and Wilson]{kapoor2024calibration}
Sanyam Kapoor, Nate Gruver, Manley Roberts, Arka Pal, Samuel Dooley, Micah Goldblum, and Andrew Wilson.
\newblock Calibration-tuning: Teaching large language models to know what they don’t know.
\newblock In \emph{Proceedings of the 1st Workshop on Uncertainty-Aware NLP}, pp.\  1--14, 2024.

\bibitem[Kirchhof et~al.(2025)Kirchhof, Kasneci, and Kasneci]{kirchhof2025position}
Michael Kirchhof, Gjergji Kasneci, and Enkelejda Kasneci.
\newblock Position: Uncertainty quantification needs reassessment for large language model agents.
\newblock In \emph{International Conference on Machine Learning}, 2025.

\bibitem[Kladny et~al.(2025)Kladny, Sch{\"o}lkopf, and Muehlebach]{kladny2025conformal}
Klaus-Rudolf Kladny, Bernhard Sch{\"o}lkopf, and Michael Muehlebach.
\newblock Conformal generative modeling with improved sample efficiency through sequential greedy filtering.
\newblock In \emph{International Conference on Learning Representations}, 2025.

\bibitem[Kuhn et~al.(2023)Kuhn, Gal, and Farquhar]{kuhn2023semantic}
Lorenz Kuhn, Yarin Gal, and Sebastian Farquhar.
\newblock Semantic uncertainty: Linguistic invariances for uncertainty estimation in natural language generation.
\newblock In \emph{International Conference on Learning Representations}, 2023.

\bibitem[Kulesza \& Taskar(2011)Kulesza and Taskar]{kulesza2011kdpp}
Alex Kulesza and Ben Taskar.
\newblock {K-DPPs: Fixed-Size Determinantal Point Processes}.
\newblock In \emph{Proceedings of the 28th International Conference on Machine Learning}, pp.\  1193–1200, 2011.

\bibitem[Kulesza \& Taskar(2012)Kulesza and Taskar]{kulesza2012determinantal}
Alex Kulesza and Ben Taskar.
\newblock Determinantal point processes for machine learning.
\newblock \emph{Foundations and Trends in Machine Learning}, 5\penalty0 (2--3):\penalty0 123--286, 2012.

\bibitem[Kwiatkowski et~al.(2019)Kwiatkowski, Palomaki, Redfield, Collins, Parikh, Alberti, Epstein, Polosukhin, Devlin, Lee, Toutanova, Jones, Kelcey, Chang, Dai, Uszkoreit, Le, and Petrov]{nq}
Tom Kwiatkowski, Jennimaria Palomaki, Olivia Redfield, Michael Collins, Ankur Parikh, Chris Alberti, Danielle Epstein, Illia Polosukhin, Jacob Devlin, Kenton Lee, Kristina Toutanova, Llion Jones, Matthew Kelcey, Ming-Wei Chang, Andrew~M. Dai, Jakob Uszkoreit, Quoc Le, and Slav Petrov.
\newblock Natural questions: A benchmark for question answering research.
\newblock \emph{Transactions of the Association for Computational Linguistics}, 7:\penalty0 452--466, 2019.
\newblock \doi{10.1162/tacl_a_00276}.
\newblock URL \url{https://aclanthology.org/Q19-1026/}.

\bibitem[Lin et~al.(2024)Lin, Trivedi, and Sun]{lin2024generating}
Zhen Lin, Shubhendu Trivedi, and Jimeng Sun.
\newblock Generating with confidence: Uncertainty quantification for black-box large language models.
\newblock In \emph{Transactions on Machine Learning Research}, 2024.

\bibitem[Loaiza-Ganem et~al.(2026)Loaiza-Ganem, Zhang, Cui, Law, and Leung]{loaiza2026conf}
Gabriel Loaiza-Ganem, Kevin Zhang, Wei Cui, Marc~T Law, and Kin~Kwan Leung.
\newblock Conf-gen: Conformal uncertainty quantification for generative models.
\newblock In \emph{International Conference on Machine Learning}, 2026.

\bibitem[Maynez et~al.(2020)Maynez, Narayan, Bohnet, and McDonald]{maynez2020faithfulness}
Joshua Maynez, Shashi Narayan, Bernd Bohnet, and Ryan McDonald.
\newblock On faithfulness and factuality in abstractive summarization.
\newblock In \emph{Proceedings of the 58th Annual Meeting of the Association for Computational Linguistics}, pp.\  1906--1919, 2020.

\bibitem[Mielke et~al.(2022)Mielke, Szlam, Dinan, and Boureau]{mielke2022reducing}
Sabrina~J Mielke, Arthur Szlam, Emily Dinan, and Y-Lan Boureau.
\newblock Reducing conversational agents’ overconfidence through linguistic calibration.
\newblock \emph{Transactions of the Association for Computational Linguistics}, 10:\penalty0 857--872, 2022.

\bibitem[Murphy(1973)]{murphy1973new}
Allan~H Murphy.
\newblock A new vector partition of the probability score.
\newblock \emph{Journal of Applied Meteorology and Climatology}, 12\penalty0 (4):\penalty0 595--600, 1973.

\bibitem[Nikitin et~al.(2024)Nikitin, Kossen, Gal, and Marttinen]{nikitin2024kernel}
Alexander Nikitin, Jannik Kossen, Yarin Gal, and Pekka Marttinen.
\newblock Kernel language entropy: Fine-grained uncertainty quantification for {LLM}s from semantic similarities.
\newblock In \emph{Advances in Neural Information Processing Systems}, 2024.

\bibitem[Papadopoulos et~al.(2002)Papadopoulos, Proedrou, Vovk, and Gammerman]{papadopoulos2002inductive}
Harris Papadopoulos, Kostas Proedrou, Volodya Vovk, and Alex Gammerman.
\newblock Inductive confidence machines for regression.
\newblock In \emph{European Conference on Machine Learning}, 2002.

\bibitem[Qiu \& Miikkulainen(2024)Qiu and Miikkulainen]{qiu2024semantic}
Xin Qiu and Risto Miikkulainen.
\newblock Semantic density: Uncertainty quantification for large language models through confidence measurement in semantic space.
\newblock In \emph{Advances in Neural Information Processing Systems}, 2024.

\bibitem[Quach et~al.(2024)Quach, Fisch, Schuster, Yala, Sohn, Jaakkola, and Barzilay]{quach2024conformal}
Victor Quach, Adam Fisch, Tal Schuster, Adam Yala, Jae~Ho Sohn, Tommi~S. Jaakkola, and Regina Barzilay.
\newblock Conformal language modeling.
\newblock In \emph{International Conference on Learning Representations}, 2024.

\bibitem[Ritter et~al.(2018)Ritter, Botev, and Barber]{ritter2018scalable}
Hippolyt Ritter, Aleksandar Botev, and David Barber.
\newblock A scalable {L}aplace approximation for neural networks.
\newblock In \emph{International Conference on Learning Representations}, 2018.

\bibitem[Ross et~al.(2025)Ross, Vouitsis, Ghomi, Hosseinzadeh, Xin, Liu, Sui, Hou, Leung, Loaiza-Ganem, and Cresswell]{ross2025textual}
Brendan~Leigh Ross, No\"el Vouitsis, Atiyeh~Ashari Ghomi, Rasa Hosseinzadeh, Ji~Xin, Zhaoyan Liu, Yi~Sui, Shiyi Hou, Kin~Kwan Leung, Gabriel Loaiza-Ganem, and Jesse~C Cresswell.
\newblock Textual {B}ayes: Quantifying uncertainty in {LLM}-based systems.
\newblock \emph{arXiv:2506.10060}, 2025.

\bibitem[Russell \& Norvig(2010)Russell and Norvig]{russel2010}
Stuart Russell and Peter Norvig.
\newblock \emph{Artificial Intelligence: A Modern Approach}.
\newblock Prentice Hall, 3 edition, 2010.

\bibitem[Shorinwa et~al.(2025)Shorinwa, Mei, Lidard, Ren, and Majumdar]{shorinwa2025survey}
Ola Shorinwa, Zhiting Mei, Justin Lidard, Allen~Z Ren, and Anirudha Majumdar.
\newblock A survey on uncertainty quantification of large language models: Taxonomy, open research challenges, and future directions.
\newblock \emph{ACM Computing Surveys}, 2025.

\bibitem[Smith et~al.(2025)Smith, Kossen, Trollope, van~der Wilk, Foster, and Rainforth]{smith2025rethinking}
Freddie~Bickford Smith, Jannik Kossen, Eleanor Trollope, Mark van~der Wilk, Adam Foster, and Tom Rainforth.
\newblock Rethinking aleatoric and epistemic uncertainty.
\newblock In \emph{International Conference on Machine Learning}, 2025.

\bibitem[Talmor et~al.(2019)Talmor, Herzig, Lourie, and Berant]{common_sense_qa}
Alon Talmor, Jonathan Herzig, Nicholas Lourie, and Jonathan Berant.
\newblock Commonsenseqa: A question answering challenge targeting commonsense knowledge.
\newblock 2019.
\newblock URL \url{https://arxiv.org/abs/1811.00937}.

\bibitem[Ulmer et~al.(2024)Ulmer, Gubri, Lee, Yun, and Oh]{ulmer2024calibrating}
Dennis Ulmer, Martin Gubri, Hwaran Lee, Sangdoo Yun, and Seong Oh.
\newblock Calibrating large language models using their generations only.
\newblock In \emph{Proceedings of the 62nd Annual Meeting of the Association for Computational Linguistics}, pp.\  15440--15459, 2024.

\bibitem[Vouitsis et~al.(2024)Vouitsis, Liu, Gorti, Villecroze, Cresswell, Yu, Loaiza-Ganem, and Volkovs]{vouitsis2024data}
No{\"e}l Vouitsis, Zhaoyan Liu, Satya~Krishna Gorti, Valentin Villecroze, Jesse~C Cresswell, Guangwei Yu, Gabriel Loaiza-Ganem, and Maksims Volkovs.
\newblock Data-efficient multimodal fusion on a single gpu.
\newblock In \emph{Proceedings of the IEEE/CVF Conference on Computer Vision and Pattern Recognition}, pp.\  27239--27251, 2024.

\bibitem[Vovk et~al.(2005)Vovk, Gammerman, and Shafer]{vovk2005algorithmic}
Vladimir Vovk, Alexander Gammerman, and Glenn Shafer.
\newblock \emph{Algorithmic learning in a random world}.
\newblock Springer, 2005.

\bibitem[Wang et~al.(2023{\natexlab{a}})Wang, Fu, Du, Gao, Huang, Liu, Chandak, Liu, {Van Katwyk}, Deac, Anandkumar, Bergen, Gomes, Ho, Kohli, Lasenby, Leskovec, Liu, Manrai, Marks, Ramsundar, Song, Sun, Tang, Veli{\v c}kovi{\'c}, Welling, Zhang, Coley, Bengio, and Zitnik]{wang2023scientific}
Hanchen Wang, Tianfan Fu, Yuanqi Du, Wenhao Gao, Kexin Huang, Ziming Liu, Payal Chandak, Shengchao Liu, Peter {Van Katwyk}, Andreea Deac, Anima Anandkumar, Karianne Bergen, \{Carla P.\} Gomes, Shirley Ho, Pushmeet Kohli, Joan Lasenby, Jure Leskovec, \{Tie Yan\} Liu, Arjun Manrai, Debora Marks, Bharath Ramsundar, Le~Song, Jimeng Sun, Jian Tang, Petar Veli{\v c}kovi{\'c}, Max Welling, Linfeng Zhang, \{Connor W.\} Coley, Yoshua Bengio, and Marinka Zitnik.
\newblock Scientific discovery in the age of artificial intelligence.
\newblock In \emph{Nature}, volume 620, pp.\  47--60, 2023{\natexlab{a}}.

\bibitem[Wang et~al.(2023{\natexlab{b}})Wang, Aitchison, and Rudolph]{wang2023lora}
Xi~Wang, Laurence Aitchison, and Maja Rudolph.
\newblock Lo{R}a ensembles for large language model fine-tuning.
\newblock \emph{arXiv:2310.00035}, 2023{\natexlab{b}}.

\bibitem[Wang et~al.(2021)Wang, Wang, Joty, and Hoi]{wang2021codet5}
Yue Wang, Weishi Wang, Shafiq Joty, and Steven~CH Hoi.
\newblock Code{T}5: Identifier-aware unified pre-trained encoder-decoder models for code understanding and generation.
\newblock In \emph{Conference on Empirical Methods in Natural Language Processing}, 2021.

\bibitem[Wei et~al.(2024)Wei, Karina, Chung, Jiao, Papay, Glaese, Schulman, and Fedus]{simpleqa}
Jason Wei, Nguyen Karina, Hyung~Won Chung, Yunxin~Joy Jiao, Spencer Papay, Amelia Glaese, John Schulman, and William Fedus.
\newblock Measuring short-form factuality in large language models.
\newblock 2024.
\newblock URL \url{https://arxiv.org/abs/2411.04368}.

\bibitem[Welbl et~al.(2017)Welbl, Liu, and Gardner]{sciq}
Johannes Welbl, Nelson~F. Liu, and Matt Gardner.
\newblock Crowdsourcing multiple choice science questions.
\newblock 2017.
\newblock URL \url{https://arxiv.org/abs/1707.06209}.

\bibitem[Xu et~al.(2024)Xu, Jain, and Kankanhalli]{xu2024hallucination}
Ziwei Xu, Sanjay Jain, and Mohan Kankanhalli.
\newblock Hallucination is inevitable: An innate limitation of large language models.
\newblock \emph{arXiv:2401.11817}, 2024.

\bibitem[Yang et~al.(2024{\natexlab{a}})Yang, Robeyns, Wang, and Aitchison]{yang2024bayesian}
Adam~X Yang, Maxime Robeyns, Xi~Wang, and Laurence Aitchison.
\newblock Bayesian low-rank adaptation for large language models.
\newblock In \emph{International Conference on Learning Representations}, 2024{\natexlab{a}}.

\bibitem[Yang et~al.(2024{\natexlab{b}})Yang, Tsai, and Yamada]{yang2024verbalized}
Daniel Yang, Yao-Hung~Hubert Tsai, and Makoto Yamada.
\newblock On verbalized confidence scores for {LLM}s.
\newblock \emph{arXiv:2412.14737}, 2024{\natexlab{b}}.

\bibitem[Zhou et~al.(2024)Zhou, Hwang, Ren, and Sap]{zhou2024relying}
Kaitlyn Zhou, Jena~D Hwang, Xiang Ren, and Maarten Sap.
\newblock Relying on the unreliable: The impact of language models' reluctance to express uncertainty.
\newblock \emph{arXiv:2401.06730}, 2024.

\end{thebibliography}
\bibliographystyle{tmlr}

\newpage
\appendix
\section{Heuristic Scores Based on Multiple Generations}\label{app:baselines}

Here we summarize the heuristic scores based on multiple generated responses that we use in \autoref{sec:experiments}. Throughout this section, we consider a query $q$ and $m$ generated responses to it, $r^{(1)}, \dots, r^{(m)}$. We use $m = 21$ in all our experiments, and use the last $m-1=20$ responses, $r^{(2)}, \dots, r^{(21)}$, to compute the scores described below. Here we only describe how these scores are calculated, we refer the reader to the corresponding references for the intuitions behind these calculations.

\textbf{Semantic Entropy} \citep{kuhn2023semantic}\textbf{.} To compute semantic entropy, the $m-1$ responses are first clustered into semantically equivalent groups; we follow \citet{kuhn2023semantic} and do this by using an entailment model \citep{he2021deberta}, which given two responses, classifies the first one as entailing, or not entailing, the second one. The entailment model is called on every pair of responses, producing a matrix $K \in \mathbb{R}^{(m-1)\times (m-1)}$, where $K_{ij}$ contains the probability that the entailment model assigns to $r^{(i+1)}$ entailing $r^{(j+1)}$ for $i,j=1,\dots, m-1$. Applying a greedy algorithm on $K$ then produces the clusters. Once these clusters are obtained, we compute the relative frequency $\hat{p}_g$ for every cluster $g$, and then semantic entropy is computed as
\begin{equation}
    -\sum_g \hat{p}_g \log \hat{p}_g.
\end{equation}

\textbf{Laplacian Uncertainties} \citep{lin2024generating}\textbf{.} We also compute two scores proposed by \citet{lin2024generating}. These scores are based on the same matrix $K$ which we computed above for semantic entropy. First, $K$ is symmetrized to obtain
\begin{equation}
    W \coloneqq \dfrac{K + K^\top}{2}.
\end{equation}
Then, the matrix $D$ is defined as a diagonal matrix, whose $i$-th diagonal entry is given by the sum of the $i$-th row of $W$. A final matrix is then computed,
\begin{equation}
    L \coloneqq I - D^{-\tfrac{1}{2}}WD^{-\tfrac{1}{2}}.
\end{equation}
From this matrix both metrics can be computed. The one we label ``Laplacian Degree'' is given by
\begin{equation}
    1 - \dfrac{\text{trace}(L)}{(m-1)^2},
\end{equation}
and the one we label ``Laplacian Eigenvalue'' is given by
\begin{equation}
    \sum_{i} \max(0, 1-\lambda_i(L)),
\end{equation}
where $\lambda_i(L)$ is the $i$-th eigenvalue of $L$.

\textbf{Kernel Entropy Uncertainty} \citep{nikitin2024kernel}\textbf{.} The last score we consider is also based on the matrix $K$. First, the matrix
\begin{equation}
    L' \coloneqq D - W
\end{equation}
is obtained, from which we compute
\begin{equation}
    A \coloneqq e^{-0.4 L'}.
\end{equation}
Then, defining $D'$ as the diagonal matrix whose $i$-th diagonal entry is $1 / \sqrt{A_{ii}}$, we compute a final matrix,
\begin{equation}
    B \coloneqq \dfrac{1}{m-1} D' A D'.
\end{equation}
The final score, which we label ``Von Neumann Entropy'', is given by
\begin{equation}
    -\sum_{i} \lambda_i(B) \log \lambda_i(B).
\end{equation}
Note that the main computational bottleneck to compute these scores is the generation of $m-1$ responses. Once this is done, computing $K$, and then each of the scores, is quick. This observation further motivates our approach: it makes sense to combine all these scores into a single confidence score, since if we have already computed one, we might as well compute the others too.

\section{Determinantal Point Processes}\label{sec:dpp}

A natural way to obtain the reference dataset from $\mathcal{D}$ is to simply sample elements uniformly at random, without replacement, to obtain $\mathcal{D}_{\text{ref}}$. However, intuitively, we would want $\mathcal{D}_{\text{ref}}$ to be as diverse as possible, so that the features we construct from it will always be as informative as possible. DPPs \citep{kulesza2012determinantal} are a way of randomly sampling diverse subsets of data. DPPs require a similarity matrix $S$, where $S_{ij}$ measures the similarity between datapoint $i$ and datapoint $j$. By leveraging this matrix, DPPs produce a random subset of the given dataset, but the resulting subset is not generated through uniform sampling without replacement; rather, it assigns higher probability to diverse (as measured by $S$) subsets. We instantiated $S$ as
\[
S_{ij} = (1 + \text{sim}(q_i, q_j))^2,
\]
and we used $k$-DPPs \citep{kulesza2011kdpp} rather than the standard DPPs, as this choice allows to specify the size of the resulting subset in advance. We follow \citet{vouitsis2024data} and use the algorithm proposed by \citet{chen2018fast}, which outputs a greedy estimate of the mode of the $k$-DPP: this can be understood as a ``maximally diverse'' subset of a specified size.

Overall, $k$-DPPs do not produce a large improvement over simply sampling the reference dataset uniformly at random. That being said, we must compute all the embeddings anyway for feature construction (which is the main computational bottleneck for using DPPs in the first place), so that the overhead of using $k$-DPPs is minimal. Since we did see a small improvement in performance from $k$-DPPs, we thus decided to use them.

\section{Additional Results}\label{app:add_results}

For completeness, we provide detailed AUROC results for all experiments across the three LLMs and all datasets and methods considered. \autoref{tab-detailed-auroc-sem4.1} reports the mean and standard deviation of AUROC over five independent random seeds for each combination with GPT4.1 generations. \autoref{tab-detailed-auroc-sem-4.1mini} displays the AUROC for each method with GPT4.1-mini generations, \autoref{tab-detailed-auroc-sem-4.1nano} displays the AUROC for each method with GPT4.1-nano generations, and \autoref{tab-detailed-auroc-llama} for LLaMA4Maverick.

\begin{table}[htbp]
\centering
\caption{Mean (standard deviation) AUROC of the different methods and datasets with \textbf{GPT4.1} generations.}
\begin{tabular}{lcccc}
\toprule
Method  & CS\_QA & NQ & SciQ & SimpleQA \\
\midrule
\pptrue & 0.743 (0.024) & 0.715 (0.008) & 0.895 (0.011) & 0.632 (0.009) \\

Logistsic Regression & 0.764 (0.023) & 0.747 (0.008) & 0.903 (0.011) & \bf 0.780 (0.023) \\
Random Forest & \bf 0.770 (0.020) & \bf 0.774 (0.008) & \bf 0.906 (0.023) & 0.779 (0.022) \\

Laplacian Degree & 0.644 (0.011) & 0.697 (0.006) & 0.564 (0.035) & 0.766 (0.016) \\
Laplacian Eigenvalue & 0.630 (0.007) & 0.708 (0.013) & 0.653 (0.011) & 0.766 (0.016) \\
Semantic Entropy & 0.614 (0.014) & 0.646 (0.009) & 0.625 (0.023) & 0.754 (0.019) \\
Von Neumann Entropy & 0.644 (0.011) & 0.698 (0.006) & 0.564 (0.035) & 0.766 (0.017) \\
\bottomrule
\end{tabular}
\label{tab-detailed-auroc-sem4.1}
\end{table}

\begin{table}[htbp]
\centering
\caption{Mean (standard deviation) AUROC of the different methods and datasets with \textbf{GPT4.1-mini} generations.}
\begin{tabular}{lcccc}
\toprule
Method  & CS\_QA & NQ & SciQ & SimpleQA \\
\midrule

\pptrue & 0.697 (0.019) & 0.703 (0.018) & 0.837 (0.033) & 0.640 (0.026) \\

Logistsic Regression & 0.736 (0.025) & 0.754 (0.008) & 0.869 (0.028) & 0.786 (0.027) \\
Random Forest & \bf 0.737 (0.030) & \bf 0.762 (0.013) & \bf 0.874 (0.027) & 0.774 (0.035) \\

Laplacian Degree & 0.646 (0.025) & 0.721 (0.005) & 0.622 (0.057) & \bf 0.788 (0.022) \\
Laplacian Eigenvalue & 0.631 (0.029) & 0.722 (0.006) & 0.699 (0.046) & 0.785 (0.024) \\
Semantic Entropy & 0.622 (0.027) & 0.691 (0.005) & 0.674 (0.031) & 0.778 (0.025) \\
Von Neumann Entropy & 0.646 (0.025) & 0.723 (0.005) & 0.622 (0.057) & \bf 0.788 (0.021) \\
\bottomrule
\end{tabular}
\label{tab-detailed-auroc-sem-4.1mini}
\end{table}

\begin{table}[htbp]
\centering
\caption{Mean (standard deviation) AUROC of the different methods and datasets with \textbf{GPT4.1-nano} generations. }
\begin{tabular}{lcccc}
\toprule
Method  & CS\_QA & NQ & SciQ & SimpleQA \\
\midrule

\pptrue & 0.626 (0.010) & 0.734 (0.015) & 0.769 (0.017) & 0.644 (0.020) \\

Logistsic Regression & 0.707 (0.009) & 0.778 (0.009) & \bf 0.838 (0.024) & \bf 0.748 (0.035) \\
Random Forest & 0.\bf 709 (0.010) & \bf 0.790 (0.011) & \bf 0.838 (0.029) & 0.745 (0.044) \\


Laplacian Degree & 0.684 (0.019) & 0.751 (0.013) & 0.668 (0.041) & 0.703 (0.033) \\
Laplacian Eigenvalue & 0.691 (0.015) & 0.741 (0.013) & 0.716 (0.038) & 0.718 (0.038) \\
Semantic Entropy & 0.660 (0.008) & 0.706 (0.014) & 0.708 (0.026) & 0.725 (0.040) \\
Von Neumann Entropy & 0.684 (0.019) & 0.750 (0.013) & 0.668 (0.041) & 0.701 (0.031) \\
\bottomrule
\end{tabular}
\label{tab-detailed-auroc-sem-4.1nano}
\end{table}





\begin{table}[H]
\centering
\caption{Mean (standard deviation) AUROC of the different methods and datasets with \textbf{LLaMA4Maverick} generations.}
\label{tab:llama4-detailed-auroc}
\begin{tabular}{lccc}
\toprule
Method & NQ & SciQ & SimpleQA \\
\midrule
\pptrue & 0.727(0.010) & \textbf{0.853(0.037)} & 0.624(0.019) \\
Logistic Regression & 0.792(0.010) & 0.817(0.078) & \textbf{0.787(0.010)} \\
Random Forest & \textbf{0.796(0.010)} & 0.842(0.038) & 0.771(0.017) \\
Laplacian Degree & 0.758(0.006) & 0.476(0.037) & 0.776(0.013) \\
Laplacian Eigenvalue & 0.762(0.008) & 0.550(0.075) & 0.768(0.006) \\
Semantic Entropy & 0.725(0.005) & 0.535(0.023) & 0.746(0.012) \\
Von Neumann Entropy & 0.758(0.006) & 0.476(0.037) & 0.778(0.014) \\
\bottomrule
\end{tabular}
\label{tab-detailed-auroc-llama}
\end{table}

\end{document}